# TensorLayer: A Versatile Library for Efficient Deep Learning Development

Hao Dong, Akara Supratak, Luo Mai, Fangde Liu, Axel Oehmichen, Simiao Yu, Yike Guo
Imperial College London
{hao.dong11,akara.supratak12,luo.mai11,fangde.liu,axelfrancois.oehmichen11,simiao.yu13,y.guo}@ic.ac.uk

## ABSTRACT

Recently we have observed emerging uses of deep learning techniques in multimedia systems. Developing a practical deep learning system is arduous and complex. It involves labor-intensive tasks for constructing sophisticated neural networks, coordinating multiple network models, and managing a large amount of training-related data. To facilitate such a development process, we propose TensorLayer which is a Python-based versatile deep learning library. TensorLayer provides high-level modules that abstract sophisticated operations towards neuron layers, network models, training data and dependent training jobs. In spite of offering simplicity, it has transparent module interfaces that allows developers to flexibly embed low-level controls within a backend engine, with the aim of supporting fine-grain tuning towards training. Real-world cluster experiment results show that TensorLayer is able to achieve competitive performance and scalability in critical deep learning tasks. TensorLayer was released in September 2016 on GitHub. Since after, it soon become one of the most popular open-sourced deep learning library used by researchers and practitioners.

## 1 INTRODUCTION

Recently we have observed the prosperity of applying deep learning into multimedia systems. Important applications include achieving visual recognition using convolution neural networks (CNN) (e.g., object recognition [23] and image generation [29]), natural language understanding using recurrent neural networks (RNN) [26] and machine strategic thinking using deep reinforcement learning (DRL) [27]. Such a prosperity has led to a booming of deep learning frameworks including TensorFlow [1], MXNet [7], Torch [9] and CNTK [30]. Developing a deep learning system typically starts with a rigorous search for an optimal neural network. A typical neural network consists of stacked neuron layers such as dropout [31], batch normalization [19], CNN and RNN. Developers rapidly evaluate varied networks by utilizing rich reference layer implementations imported from open source libraries including Caffe [20], Keras [8], Theano [6], Sonnet [11] and TFLearn [10].

Deep learning systems are increasingly interactive [3]. This has led to three transitions in their development landscape. Firstly, datasets are becoming dynamic. The emergence of learning systems that involve feedback loops, e.g., DRL and active learning [13], has spawn numerous requests for manipulating, consolidating and querying datasets. Secondly, models are becoming composite. To achieve precise recognition, a deep neural network, e.g., dynamic neural networks [4] and generative adversarial networks (GANs) [29], can need to activate varied neuron layers according to input features. Thirdly, training is becoming continuous. Samples, features, human insights, and operation experiences can be produced even after deployment. It thus becomes necessary to constantly optimize the hyper-parameters of a neural network by supporting human-in-the-loop.

The growing interactivity complicates deep learning development. Developers have to spend many cycles on integrating components for experimenting neural networks, managing intermediate training states, organizing training-related data, and enabling hyper-parameter tuning in responses to varied events. To reduce required cycles, we argue for an integrative development approach where the complex operations towards neural networks, states, data, and hyper-parameters are abstracted and provided within complementary modules. This spawns an unified environment where developers are able to efficiently explore ideas through high-level module operations, and apply customizations to modules only if necessary. This approach does not intend to create module lock-in. Instead, modules are modelled minimal single-function blocks that share an interaction interface, which allows easy plug-ins of user-defined modules.

TensorLayer is a community effort to achieve this vision. It is a modular Python library that provides easy-to-use modules to facilitate researchers and engineers in developing complex deep learning systems. Currently, it has (1) a *layer* module that provides reference implementation of neuron layers which can be flexibly interconnected to architect neural networks, (2) a *model* module that can help manage the intermediate states incurred throughout a model life-cycle, (3) a *dataset* module that manages training data which can be used by both offline and online learning systems, and (4) a *workflow* module that supports asynchronous scheduling and failure recovery for concurrent training jobs. In addition to using high-level module APIs, TensorLayer users are allowed to offload low-level functions into execution backends to achieve fine-grain controls towards a training process. This transparent interface design is particular useful when addressing domain-specific problems.

TensorLayer implementation is optimized for performance and scalability. It adopts TensorFlow as the distributed training and inference engine. Delegation into TensorFlow exhibits negligible overhead. In addition, TensorLayer uses MongoDB as the storage backend. This backend is augmented with an efficient stream controller for managing unbounded training data. To enable automation, this controller is able to batch results from a dataset query and spawns batch training tasks accordingly. For efficiently handling large data objects like videos, TensorLayer uses GridFS as a blob backend, and makes MongoDB act as an sample indexer. Finally, TensorLayer implements an agent pub-sub system to achieve asynchronous training workflow. Agents can be placed onto various kinds of machines and subscribe to independent tasks queues. These queues are managed by in reliable storages so that failed tasks can be replayed automatically.

In contrast to other TensorFlow-based libraries such as Keras and TFLearn, TensorLayer permits straightforward low-level controls within the execution of layers and neural networks. In addition, it provides extra dataset and workflow modules to free users from labor-intensive data pre-processing, post-processing, module serving and data management tasks. Last but not least, TensorLayer does not create library lock-in. Its unified module interaction interface can accept layers and networks imported from Keras and TFLearn in a non-invasive manner.

TensorLayer was released on Github [1] in September 2016, and soon become one of the most popular open-sourced deep learning libraries [14]. By July 2017, it has received more than 1900 stars and has formed an active development community. We demonstrate its effectiveness through real-world applications, where TensorLayer are used to implement DRL, GANs, model cross-validation and hyper parameter optimization [5]. These applications were previously challenging to develop and requires expensive development effort for integrating storage, fault-tolerance, computations and cluster management components. The use of TensorLayer significantly accelerates such a process.

## 2 ARCHITECTURE

We describe the architecture of TensorLayer in Figure 1. A deep learning developer writes a multimedia system using helper functions from TensorLayer . These functions range from providing and importing layer implementations, to building neural networks, to managing states involved throughout model life-cycles, to creating online or offline datasets, and to writing a parallel training plan. These functions are grouped into four modules: layer, network, dataset, and workflow. In the following, we describe these modules, respectively.

[1] https://github.com/zsdonghao/tensorlayer

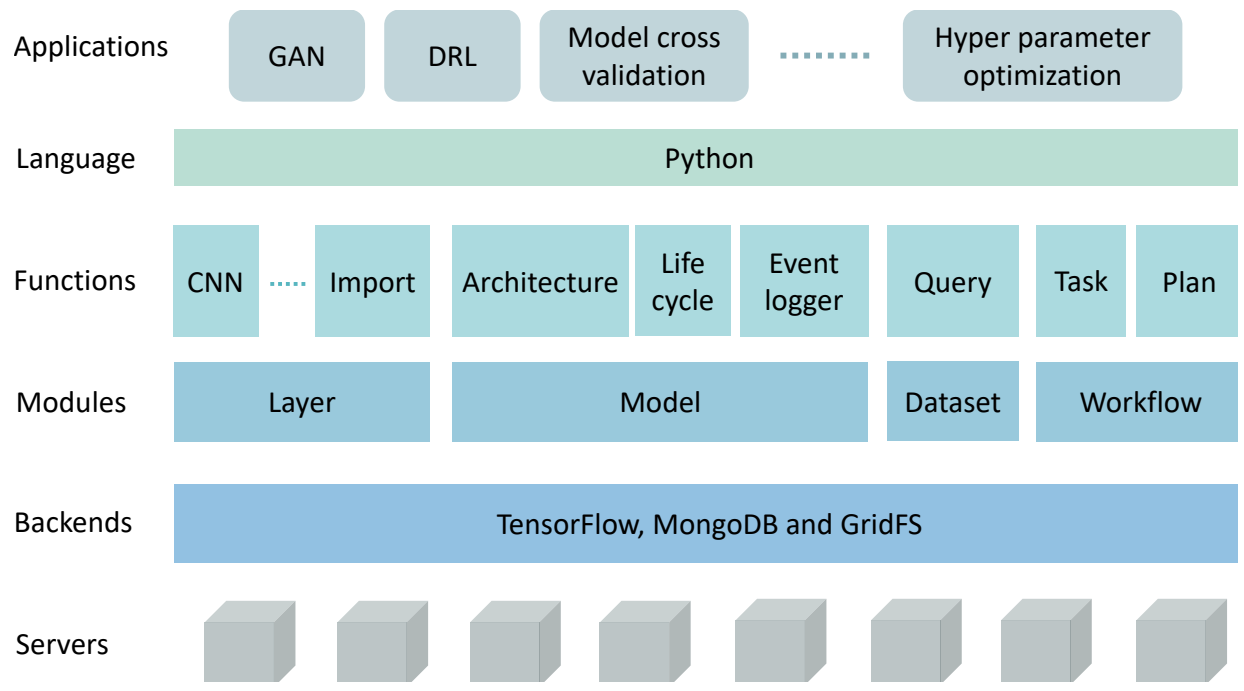

**Figure 1: TensorLayer architecture.**

**Table 1: TensorLayer (TL) and TensorFlow (TF) benchmark.**

| | CIFAR-10 | PTB LSTM | Word2vec |
|---|---|---|---|
| TL | 2528 images/s | 18063 words/s | 58167 words/s |
| TF | 2530 images/s | 18075 words/s | 58181 words/s |

### 2.1 Layer Module

Layers are the core bricks of a neural network. TensorLayer provides a layer module that includes reference implementations of numerous layers, such as CNN, RNN, dropout, dropconnect, batch normalization and many others. Layers are stacked to create a neural network with a declarative fashion, similar to the extensively used Lasagne [12]. Each layer is given an unique key for helping developers achieve parameter sharing. The networks are delegated to TensorFlow. TensorLayer inherits from TensorFlow to run on hybrid and distributed platforms. A concern towards TensorLayer is performance overhead. We investigate this by running classic models [32] using TensorLayer and native TensorFlow implementations on a Titan X Pascal GPU. Table 1 shows that TensorLayer exhibits negligible overhead in all models.

Layers can be flexibly composed and customized. They can be embedded with control functions that are lazily evaluated within TensorFlow to adjust training behaviours. This transparency design is favoured by TensorLayer users, in particular when they are addressing domain-specific problems that require carefully customized models. In addition, to ease migration, TensorLayer allows importing external layer and network implementations from other TensorFlow wrappers such as Keras, TFLearn and TFSlim by using the LambdaLayer.

### 2.2 Model Module

Model is the logical representation of a self-contained functional unit, and can be trained, evaluated and deployed in production. Each model has an unique network structure. During training, the model can have different versions or states (i.e., weights). States can be persisted, cached and reloaded. We use MongoDB as the storage backend. Compared to other storage providers such as HBase or MySQL, MongoDB is out-of-box deployable, simple to use, and rich in third-party management tools. In terms of performance, it is able to achieve sufficient throughput for processing tensors which are the dominant data type in deep learning.

TensorLayer supports recording user-defined model events. Conventional events reflect training steps, learning speed, and accuracy. They are often used for diagnosing a training process in order to achieve, for example, model versioning [25] and interactive learning [21].

## 2.3 Dataset Module

The dataset module is used for managing training samples and prediction results. They are stored as documents in MongoDB. Each document contains a unique key, sample, label and user-defined tags. Datasets are specified by declarative queries that carry conditions towards tag fields. Queries create views of underlying data, and thus do not cost extra storage.

Data are modelled as general streaming datasets. Each dataset is given a *stream* controller that constantly monitor the availability of samples and predictions, and then trigger corresponding training tasks towards this dataset. Intermediate training states are cached in memory and later reloaded among batches. The streaming abstraction is able to unify the declaration of offline and online data. To speed up training efficiency, changes to a dataset is batched until become significant.

TensorLayer optimizes dataset performance thoroughly. Firstly, TensorLayer creates indexes for frequently visited tags to accelerate row selection. Secondly, datasets can be cached locally and partitioned to distribute workloads. Thirdly, chunky data are compressed and sent in batches to improve I/O efficiency. Thirdly, in addressing big blobs such as videos, TensorLayer adopts GridFS as the blob store. In such a case, the rows in MongoDB carry pointers to the locations of sample blobs in GridFS. This two-layer storage architecture is transparent to developers.

## 2.4 Workflow Module

The workflow module provides task abstraction to enable fault-tolerant asynchronous training. A training task is uniquely identified by 3-tuple: an input dataset key, a model key, and an output dataset key. It is inserted into a task queue subscribed by an agent. An agent can perform CPU / GPU training task or any user-defined function written in Python. Task completion messages are published onto a notification queue subscribed by an agent master. This pub-sub system is naturally asynchronous. Multiple tasks can form a training plan to be scheduled by the master. Queues are persisted in the storage backend to ensure that failed tasks are replayed.

The workflow module simplifies the implementation of model group operations and learning systems that involves asynchronous feedback loops in operation. It is also useful for complex cognitive systems that have training dependency among components. For example, the developer of an image captioning system [28] first trained a CNN to understand the context of images, and then trained a RNN decoder to generate description based on recognized context. This example thus forms a two-stage asynchronous training plan that can be supported by TensorLayer .

# 3 APPLICATIONS

This section presents a comprehensive study of deep learning applications that can benefit from using TensorLayer in terms of development efficiency. Relevant source code is publicly available on Github [2].

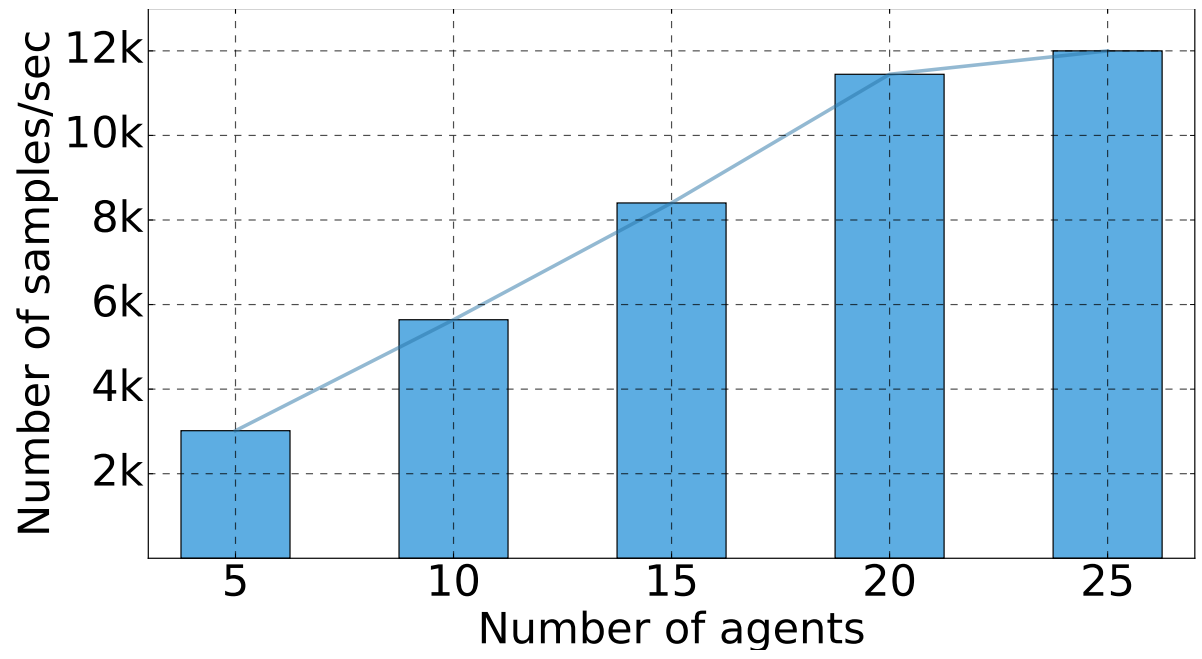

**Figure 2: Training throughput vs. number of agents used for generating training samples.**

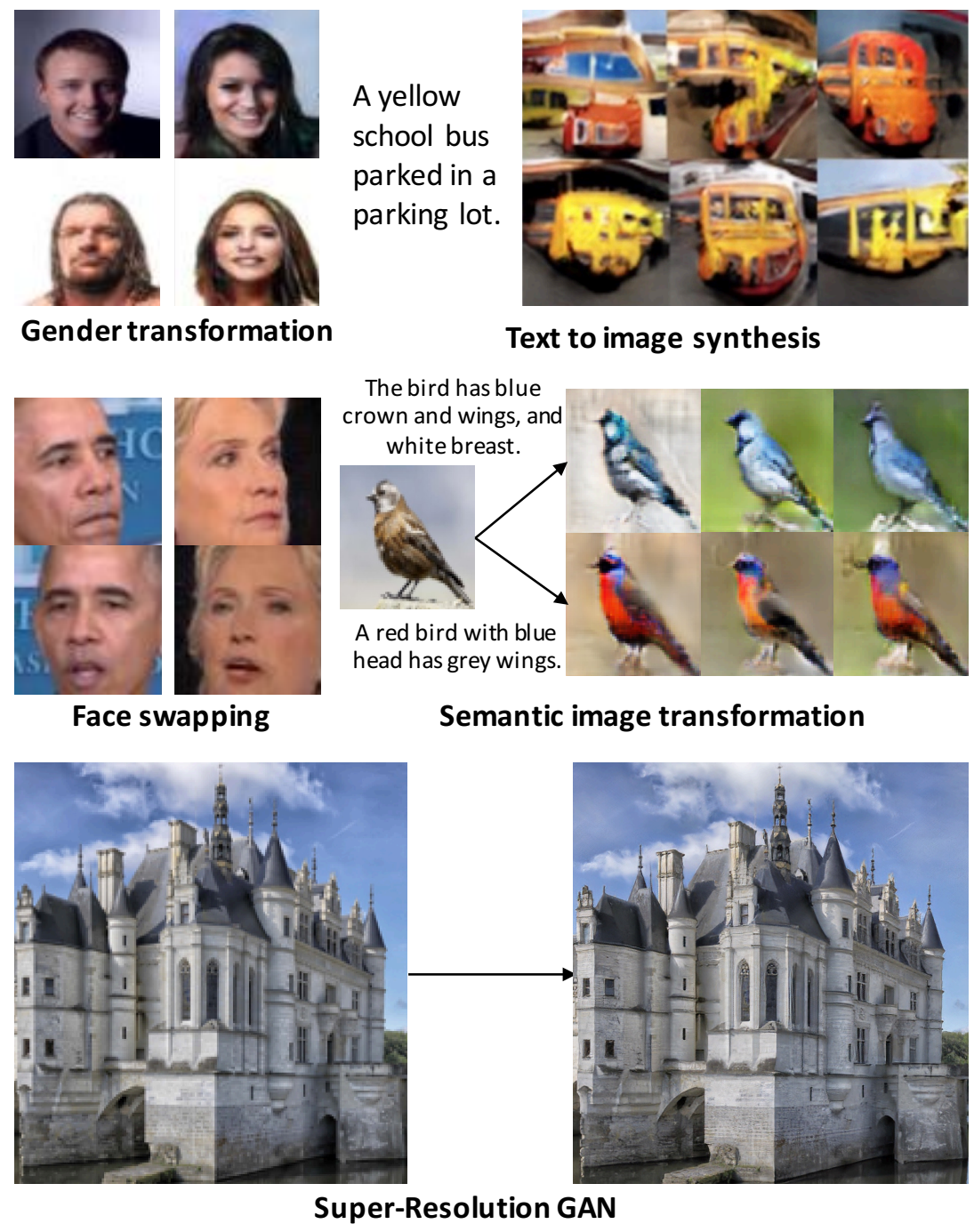

**Figure 3: Highlighted TensorLayer applications.**

**Generative adversarial networks.** GANs have become a popular deep learning based generative framework for multimedia tasks. The discriminator network of GANs has two source inputs, which is different from common deep learning architectures. TensorLayer enables developers to efficiently construct network architectures of GANs, and control the dynamics of a training process, and achieve parameter optimization. We take DCGAN [29], an image generation network, as an instance to evaluate the helpfulness of TensorLayer . While achieving identical training efficiency, TensorLayer implementation has 187 lines of code (LOC), which is 75% smaller than the published TensorFlow implementation (746 LOC). We use Super-Resolution GAN (SRGAN) [24] as another example. The TensorLayer-based implementation of SRGAN is 526 LOC in length. This is smaller than many other open-sourced implementations that often have more than a thousand LOC.

[2] https://github.com/akaraspt/tl_paper

**Deep reinforcement learning.** A DRL application is a great example that showcases a joint usage of the layer, model, dataset and workflow modules. Specifically, developers can use TensorLayer to build a DRL model, manage model's states between iterations, and create training data that will be constantly generated by concurrent game players. To demonstrate this, we implement a distributed asynchronous DRL system in a cluster that has 10 Gbps connectivity. The system trains an agent for playing Atari pong game [22] on a GTX 980 GPU. The trainer keeps receiving samples (i.e., observations, actions and rewards) from game players simulated by TensorLayer agents. Trained network models are shared with all players via the model module of TensorLayer . Figure 2 illustrates the scalability of TensorLayer in powering such a system. The training throughput is linearly increasing with more joining agents, until it reaches the maximum capacity of the GPU.

**Hyper parameter optimization and cross-validation.** These two machine learning jobs are necessary for addressing domain-specific problems, e.g., medical signal processing [2], which usually do not have universally effective models. Hence, they help developers explore various models and evaluate their performance. Previously, they were implemented using ad-hoc components. These implementations incurred high maintenance cost, and reduced task efficiency due to the cross-component overhead (e.g., serialization and network transfer). Integrating them with TensorLayer significantly reduces the development and maintenance complexity. In addition, experiment results show that TensorLayer can gently increase task parallelism while only incurring low memory and I/O overhead within the shared data infrastructure.

**More applications.** There are many more applications that have benefited from using TensorLayer . We highlight a few of them here: multi-model research [16], image transformation [17, 18], and medical signal processing [2, 15]. Their results are illustrated in Figure 3.

## 4 AVAILABILITY

TensorLayer is open sourced under the license of Apache 2.0. It can be used in Linux, Mac OS and Windows environments. TensorLayer has a low adoption barrier. It provides a multilingual documentation, massive tutorials and thorough examples, such as CNNs (like VGG, ResNet, Inception), text-related applications (like text generation, Word2vec, machine translation, image captioning), GANs (text-to-image synthesis, CycleGAN, stackGAN, SRGAN), reinforcement learning algorithms (like Deep Q-Network, Policy Gradients, Asynchronous Advantage Actor-Critic (A3C)) and etc.

## 5 ACKNOWLEDGE

We would like to thank TensorLayer Open Source Community for numerous fruitful discussions. We also appreciate the feedbacks from Biye Jiang affiliated with University of California Berkeley. Hao Dong is supported by the OPTIMISE Portal. Akara Supratak is supported by the Faculty of Information and Communication Technology, Mahidol University (MUICT). Luo Mai is supported by a Google Fellowship in Cloud Computing. Axel Oehmichen is supported by the Innovation Medicines Initiative (IMI) eTRIKS project. Simiao Yu is supported by Jaywing plc.